\pdfoutput=1

\documentclass[letterpaper, 10 pt, conference]{ieeeconf}  %

\IEEEoverridecommandlockouts                              %

\usepackage{graphics} %
\usepackage{calc}
\usepackage{epsfig} %
\usepackage{times} %
\usepackage{amsmath} %
\usepackage{amssymb}  %
\usepackage[dvipsnames]{xcolor}
\usepackage[most]{tcolorbox}
\let\labelindent\relax %
\usepackage{enumitem}  %

\newtcolorbox{promptbox}[1]{
    colback=white,           %
    colframe=black!75,       %
    fonttitle=\bfseries\sffamily\fontsize{8}{10}\selectfont,
    title=#1,
    arc=0mm,                 %
    outer arc=0mm,
    left=10pt, right=10pt, top=8pt, bottom=8pt,
    enhanced,
    attach boxed title to top left={xshift=3mm, yshift=-2mm},
    boxed title style={colback=black!75, sharp corners}
}

\newcommand{\prompttext}[1]{{\sffamily\bfseries\fontsize{8}{10}\selectfont #1}}

\usepackage{siunitx}
\usepackage{makecell}
\usepackage{tabularx}
\newcolumntype{C}{>{\centering\arraybackslash}X}
\usepackage{subfig}
\usepackage[para,online,flushleft]{threeparttable}
\usepackage{soul}
\usepackage{bm}
\usepackage[font=small, labelfont=bf]{caption}
\usepackage{hyperref}
\hypersetup{colorlinks=true, urlcolor=blue, citecolor=magenta, linkcolor=black}

\usepackage{dblfloatfix}

\usepackage[
 doi=false,isbn=false,url=false,eprint=false,
 backend=biber,
 style=ieee,
 bibstyle=ieee, %
 citestyle=numeric-comp, %
 sortcites=true, 
 giveninits=true, %
 backref=false]{biblatex}
\AtEveryBibitem{%
  \clearfield{volume}%
  \clearfield{number}%
  \clearfield{pages}%
}
\usepackage{balance}

\DeclareRobustCommand{\citet}[1]{\citeauthor{#1}~\cite{#1}}

\title{\LARGE \bf Synergistic Action and Style Priors for Learning Legged Locomotion}
\author{Oliver Hausdörfer$^{1}$, Alexander von Rohr$^{1}$, Éric Lefort$^{2}$, and Angela P. Schoellig$^{1}$%
\thanks{$^{1}$The authors are with the Learning Systems and Robotics Lab, Technical University of Munich; and Munich Institute of Robotics and Machine Intelligence. This work was supported by the Robotics Institute Germany under BMBF grant 16ME0997K and by the Humboldt Professorship for Robotics and Artificial Intelligence. $^{2}$É. Lefort is with the University of Toronto. Research conducted while at LSY, TUM. Email: {\tt\small oliver.hausdoerfer@tum.de}.}%
}

\begin{document}
\makeatletter
\let\@oldmaketitle\@maketitle%
\renewcommand{\@maketitle}{\@oldmaketitle%
    \centering
    \vspace*{1mm}
    \includegraphics[width=1.0\textwidth]{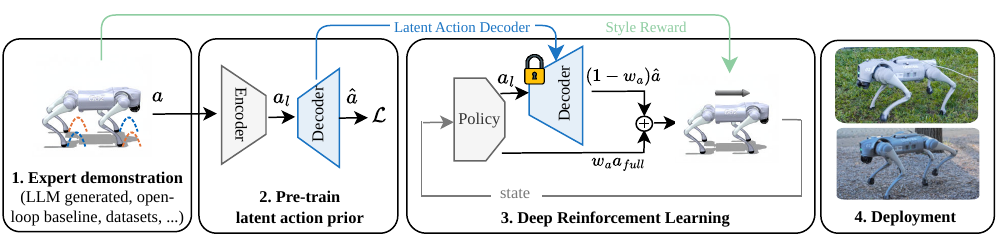}
    \captionof{figure}{
    We propose to combine a simple action space prior with a single imitation reward term --- both obtained from a small expert demonstration dataset. The action prior is an autoencoder-style pretrained decoder that exploits correlations between the expert actions in the demonstration. During policy learning, the agent outputs latent actions $a_l$ and a residual in the full action space $a_{\mathrm{full}}$ --- the decoder remains frozen. The resulting policies can be deployed on hardware for position and direct \SI{500}{\hertz} torque control.}

    \label{fig:method}
    \vspace*{-3mm}
}
\makeatother
\maketitle
\setcounter{figure}{1}

\thispagestyle{empty}
\pagestyle{empty}

\begin{abstract}
Complex reward tuning remains a problem for learning locomotion behaviors, particularly at the torque level. Demonstration-guided imitation rewards and action-space priors mitigate this burden, but each introduces its own tuning effort or assumes demonstration data that is costly to obtain. We propose to leverage a simple action space prior in combination with a single imitation reward term to learn locomotion, both derived from a small state--action demonstration set. We find that the synergy between the priors is complementary: the action prior restricts exploration to the demonstrated action manifold, while the imitation term supplies the temporal structure required for imitation. This synergy yields faster convergence, higher task reward, and reduced sensitivity to the introduced hyperparameters than either prior in isolation. Addressing the question of the required demonstrations, we further show that a large language model (LLM) can design the underlying feedforward gait controllers through a simulation-in-the-loop procedure, removing the need for motion capture or manual trajectory design. The resulting policies transfer to a Unitree Go2 for both position control and direct \SI{500}{\hertz} torque control, demonstrating an end-to-end pipeline from LLM output to low-level robot control. Videos and code are available at \url{https://sites.google.com/view/latent-action-priors}.
\end{abstract}

\section{Introduction}
Deep reinforcement learning (DRL) has successfully enabled robots to learn complex behaviors from interactions with their environment~\cite{Lee.2020,he2025asap}. When learning from scratch, often 15 or more regularizing reward terms, including feet contact schedules, are needed to shape the desired learning outcome~\cite{bohlinger2024policyrunallendtoend}. Otherwise, the obtained motions look unnatural and cannot be deployed to hardware.  This problem is even more pronounced for direct torque control, which lacks the structural priors of position-based action spaces. Torque control is nevertheless desirable, as it avoids PD-gain tuning and enables compliant, reactive contact.

Reward tuning can partially be circumvented by leveraging reference motions and demonstrations. Popularly, this is done through guiding learning using imitation reward terms~\cite{Peng.2018}. Imitation rewards necessitate a reference that is defined through a sequence of states. Such reference for quadruped can be obtained through hand-crafted motions in animation software~\cite{zargarbashi2024robotkeyframinglearninglocomotionhighlevel} or reference dog motions recorded via motion capture~\cite{Peng.2018}. Notably, the required motions are state-only and lack action labels.

In this work, we propose to combine imitation rewards with a simple prior in the action space. Our hypothesis is that an informed action space biases exploration toward the actions observed in the demonstration, thereby reinforcing the behavior that the style reward independently rewards. We show that this combination provides a strong prior even under challenging direct torque control. The synergy of both priors further enables robustness with regard to the introduced hyperparameters, and improved sample efficiency during DRL training, i.e., faster convergence. Whereas prior works rely on hand-designed references to obtain the required state–action demonstrations, we use LLM-based code generation to automate this manual design step.

\textbf{Our experiments are organized around four questions}. \textbf{i)} Does our action space prior provide a suitable signal for learning locomotion? \textbf{ii)}~Do action and style priors interact synergistically during policy learning on standard locomotion benchmarks? \textbf{iii)}~Can an LLM supply the required state--action demonstrations?, and \textbf{iv)}~Do the resulting policies transfer to hardware under both position and torque control?

\textbf{In summary, our main contributions are}:
\begin{itemize}
    \item We show that combining a simple data-driven action prior with an imitation reward term yields synergies that enable efficient locomotion learning with robustness towards hyperparameter choices, which neither prior achieves in isolation.
    \item We demonstrate that LLMs can effectively design reference controllers --- using established methods --- to generate the required state-action demonstrations, and that the resulting policies are deployable on hardware for position and low-level torque control.
\end{itemize}

\section{Related Work}
\textbf{Locomotion Learning.} DRL has become the predominant method to learn locomotion~\cite{Lee.2020, rudin2022learningwalkminutesusing}. Typically, the policies output joint position targets that are tracked using a lower-level PD controller~\cite{, bohlinger2025gaiteightefficientonrobot, bohlinger2024policyrunallendtoend}. The joint position targets can either be absolute joint values or relative to a reference position. The latter helps to bias the policy towards a natural gait when, for instance, a standing pose is chosen as reference. Other works design an entire reference feedforward trajectory for the foot positions~\cite{Lee.2020}.
The DRL policy then learns a residual around this reference to enable robustness. A similar effect is achieved by learning the parameters of (coupled) oscillators, where the
phase output provides smooth trajectories for joint position
target~\cite{Bellegarda.2022}. Typically, aforementioned methods rely on manually designed references and require up to 15 regularizing reward terms that need to be meticulously tuned.

Other works learn policies directly in torque-control mode~\cite{Sood2023DecAPD, chen2023learningtorquecontrolquadrupedal, Soni_2023}. Despite its reported advantages in robustness and responsiveness, torque-control learning remains comparatively less explored due to the additional challenges it presents in learning and deployment. Designing reference trajectories and biases for torques is more difficult than for positions, which further increases engineering difficulties.

In contrast, we consider a data-driven action prior learned from state-action demonstrations as an inductive bias for learning low-level control commands.

\textbf{Learning from Demonstrations.} Sources of expert demonstrations that provide expert states include motion capture data~\cite{Peng.2018} and manually designed trajectories~\cite{Lee.2020}.  DeepMimic~\cite{Peng.2018} leverages expert states to formulate an imitation reward, which works well for the low data regime \cite{Wang.2024, GrandiaRuben.2024}. Adversarial approaches, such as adversarial motion priors (AMP)~\cite{Peng_2021}, employ generative adversarial training methods to distinguish between expert transitions and learned transitions~\cite{Ho.2016}. Adversarial methods are generally known to be unstable, and thus introduce significantly longer training times and additional hyperparameters compared to PPO~\cite{adv}. 

In case sufficient expert data is available, one can use Imitation Learning directly~\cite{survey}, such as behavioral cloning (BC) \cite{MichaelBain.1995}. Offline-to-online DRL leverages prior data by learning a value function~\cite{smith2023transfer, nair2020awac}, which requires the data to cover the states and actions the policy visits.

Demonstrations are also used for residual policy learning~\cite{Luo.2024b}. The reference can either be a simple baseline, such as a trotting gait cycle~\cite{Luo.2024} or more complex motion trajectories~\cite{li2024fld}.
Priors in the action space have been shown to be effective for both manipulation and locomotion tasks~\cite{Biza2021}. In our work, we combine action priors with a style reward and demonstrate synergies for locomotion learning in the little-data regime.

\begin{figure*}[!t]
  \centering%
  \subfloat[]{\includegraphics[width=0.156\textwidth]{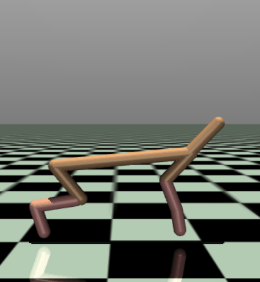}}%
  \hspace{0.012\textwidth}%
  \subfloat[]{\includegraphics[width=0.156\textwidth]{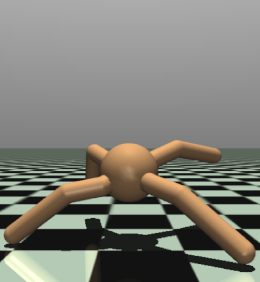}}%
  \hspace{0.012\textwidth}%
  \subfloat[]{\includegraphics[width=0.156\textwidth]{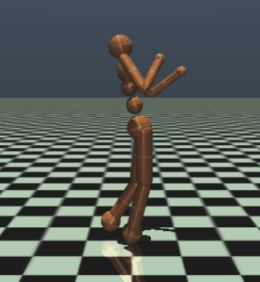}}
  \hspace{0.012\textwidth}%
  \subfloat[]{\includegraphics[width=0.156\textwidth]{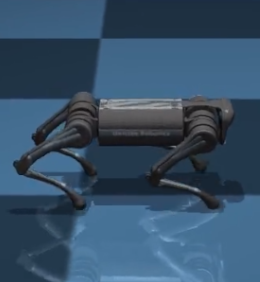}}
  \hspace{0.012\textwidth}%
  \subfloat[]{\includegraphics[width=0.156\textwidth]{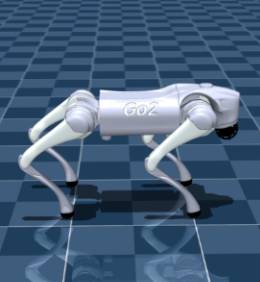}}
  \hspace{0.012\textwidth}%
  \subfloat[]{\includegraphics[width=0.156\textwidth]{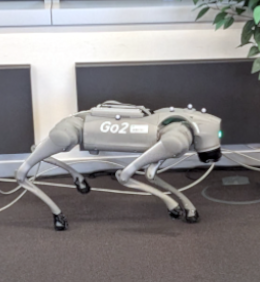}}
  \caption{We evaluate the synergies of combining the simple data-driven action prior with a style reward term on (a) - (c) three Gymnasium Mujoco \cite{towers2024gymnasium} environments of varying difficulty, and (d) a simulated Unitree A1 quadruped with position and torque control from the loco-mujoco benchmark~\cite{AlHafez.2023}. (e) - (f) We then test our approach on a Unitree Go2 for two gait types and terrains using LLM-generated demonstrations, and deploy the policy on hardware for both position and direct \SI{500}{\hertz} torque control.}%
  \label{fig:agents}%
\end{figure*}%

\textbf{Action Representations.} Many natural movements can be described using a lower-dimensional space. Such motion synergies have been exploited to learn whole-body motion for humanoids~\cite{Luo.2024b} or dexterous robotic manipulation with hands \cite{Rivera.2021, Zeng.2018}. Those works either manually design action representations or discover them during learning. An autoencoder-style action prior learned from expert data is introduced by \citet{Luo.2024b} for a humanoids. \citet{Luo.2024b} shows that various motions can be reconstructed, however, it tests in simulation, assumes a given motion capture dataset, and designs a variety of reward terms in addition to requiring the demonstration. We show that a similar autoencoder-style architecture can be leveraged to learn both joint position and low-level torque actions.

\section{Method}\label{sec:method}

We consider DRL for locomotion learning with access to demonstrations containing states and actions. The data is used in two complementary ways: first, to learn a latent action prior that biases the exploration toward actions observed in the demonstrations, and second, to define a single imitation reward term that encourages the learned behavior to remain close to the demonstrated motion (\autoref{fig:method}).

\textbf{Problem Formulation.} We consider a Markov decision process with state $s_t \in \mathcal{S}$, action $a_t \in \mathcal{A}$, transition dynamics $p(s_{t+1}\mid s_t, a_t)$, and a stochastic policy $\pi_\theta(a_t \mid o_t)$ parameterized by $\theta$, where $o_t$ denotes the policy observation at time step $t$.
The policy is optimized with respect to a reward function $r_t$, and the objective is to maximize the expected discounted return $J(\pi_\theta) = \mathbb{E}\left[\sum_{t=0}^{\infty} \gamma^t r_t \right].$

In addition to environmental interaction, we assume access to one or more demonstration trajectories
\begin{equation}
    \mathcal{D} = \{\tau_i\}_{i=1}^{N}, \qquad
    \tau_i = \{(s_{t}^{\mathrm{exp}}, a_{t}^{\mathrm{exp}})\}_{t=1}^{T_i},
\end{equation}
consisting of expert states $s_{t}^{\mathrm{exp}}$ and expert actions $a_{t}^{\mathrm{exp}}$. %

\textbf{Latent Action Pretraining.} To guide exploration toward actions that are consistent with the demonstrations, we learn a prior over the action space from the demonstrated actions, and we demonstrate our method for both bases. The actions can be joint torques or joint positions. We train an encoder-decoder model on the expert actions in $\mathcal{D}$, where the encoder maps an expert action $a^{\mathrm{exp}}$ to a latent representation and the decoder reconstructs the action, with
\begin{equation}
    a_l = E(a^{\mathrm{exp}}), \qquad
    \hat{a}^{\mathrm{exp}} = D(a_l).
\end{equation}
The latent action $a_l \in \mathbb{R}^{d_l}$ provides a lower-dimensional representation of the demonstrated action, where $d_l \leq \dim(\mathcal{A})$.

We train the encoder and decoder by minimizing a reconstruction objective over the demonstration dataset,
\begin{equation}
    \mathcal{L}_{\mathrm{rec}}
    =
    \mathbb{E}_{a^{\mathrm{exp}} \sim \mathcal{D}}
    \left[
        \left\|
            D(E(a^{\mathrm{exp}})) - a^{\mathrm{exp}}
        \right\|_2^2
    \right].
\end{equation}
To enable compatibility with standard bounded policy outputs for policy learning, we additionally regularize the latent representation to remain within a fixed bound:
\begin{equation}
    \mathcal{L}_{\mathrm{prior}}
    =
    \mathcal{L}_{\mathrm{rec}}
    +
    \lambda_l
    \,
    \mathbb{E}_{a^{\mathrm{exp}} \sim \mathcal{D}}
    \left[
        \ell_{\mathrm{bound}}(E(a^{\mathrm{exp}}))
    \right],
\end{equation}
where $\lambda_l$ weights the regularization term and $\ell_{\mathrm{bound}}(\cdot)$ penalizes latent values outside the range of $[-1,1]$.

During reinforcement learning, the policy outputs both a latent action $a_l$ and a residual action $a_r$ in the full action space. The executed action is given by
\begin{equation}
    a = (1 - w_a) D(a_l) + w_a a_r,
\end{equation}
where $w_a \in [0,1]$ controls the contribution of the residual action. This parameterization biases the policy toward actions by the demonstrations while retaining the ability to deviate from them when necessary. In particular, the residual enables recovery from off-demonstration states and perturbations.

In practice, both encoder and decoder are parameterized as single-hidden-layer multilayer perceptrons with hidden size $2\,\dim(a_l)$ and \texttt{tanh} activations. This supervised learning problem on a small dataset incurs negligible additional cost --- a few minutes of training on a workstation --- compared to DRL training.

\textbf{Rewards.} The overall reward is given by the task reward formulation $r_{task}$ and a single style reward term $r_{style}$. This formulation encourages the policy to optimize task performance while remaining
close to the demonstrated motion:
\begin{equation}
    r = w_{\mathrm{task}} r_{task} + w_{\mathrm{style}} r_{style},
    \label{eq:rew}
\end{equation}
where we set $w_{task}=0.67$ and $w_{style}=0.33$ similar to~\cite{Peng.2018}. 

The imitation style reward encourages the agent to exhibit similar joint positions to the expert demonstration:

\begin{equation}
    r_{style}= \exp\left[-\left(\sum_{j}\lVert q_{exp}^j-q^j \lVert_2^2\right)\right].
    \label{eq:style}
\end{equation}

This reward encourages the agent to match the joint positions of the expert at each time step and is computed as the difference between the $j$ joint positions of the agent $q^j$ and the expert demonstration $q^j_{exp}$. The style reward requires the use of a phase variable $\phi \in [0, 2 \pi]$ to match the style from the cyclic demonstrations to the entire episode and ensure a Markovian reward. The policy receives the sine and cosine of this phase as inputs. Note that the phase variable is not required for the action prior.

For the Gymnasium environments~\cite{towers2024gymnasium}, we use the default task rewards of the environments. For the quadrupeds, the task reward is to follow a user-commanded velocity and is computed using the exponential kernel from~\citet{rudin2022learningwalkminutesusing}.

\textbf{Policy Optimization.} We optimize the policy parameters~$\theta$ with deep reinforcement learning using the previously described action parameterization and reward to maximize the expected discounted return. We use proximal policy optimization (PPO)~\cite{schulmann.2017} to solve the resulting reinforcement learning problem, although our approach is compatible with other frameworks as well. PPO is applied directly to the policy parameterization described above, such that policy updates jointly adapt the latent-action output and the residual-action output while the action prior remains a fixed inductive bias.

\textbf{LLM-aided Demonstration Generation.}\label{ssec:llm} We rely on the data provided by the benchmarks for the benchmark problems (\autoref{sec:exp_details}). In practice, however, obtaining demonstrations that contain states and actions is a bottleneck. Thus, to obtain the required demonstrations for the Go2, a large language model (LLM) designs the underlying feedforward gait controllers through a simulation-in-the-loop procedure. The idea is to iterate on the design of simple feed-forward controllers in simulation, which is then rolled out to obtain the demonstration dataset. The feedforward gait generation produced by the LLM is based on established methods using a scripted swing-stance foot-trajectory generator~\cite{zeng2019legtrajectory} closed by analytical inverse kinematics~\cite{iscen2018pmtg}. The controllers generate foot trajectories and compute corresponding joint trajectories using inverse kinematics. The joint targets are tracked by low-level PD controllers. The LLM then adjusts the exposed parameters.

The LLM is presented with constraints to solve the task (\autoref{fig:prompt_example}). The prompt includes the \prompttext{Task} (generating a feedforward controller for the Unitree Go2 in a specific gait), \prompttext{Requirements} (gait styles and actuator gains), and \prompttext{Key Parameters} that should be varied. The simulation-in-the-loop step also acts as a filter on infeasible or hallucinated parameters: gait frequencies, step lengths, or foot placements outside the robot's kinematic and actuation limits produce visibly failing rollouts and contact patterns and are rejected before any data is recorded. Only controllers that satisfy the stated requirements are rolled out for a few gait cycles to obtain the demonstration dataset $\mathcal{D}$ consisting of states and actions. Selection is semi-automatic: the LLM checks the requirements via the printed metrics, and a human accepts the final controller after visual inspection.

\begin{figure}[tbp]
\centering
\begin{promptbox}{LLM Prompt: Gait Generation}
    \fontsize{8}{9}\selectfont
    \ttfamily %
    
    \textbf{Task:} Create a feedforward gait generator for the Unitree Go2. Iterate until the requirement are met.\\

    \textbf{Requirements:}
    \begin{itemize}[leftmargin=1.5em, nosep]
        \item Generate foot positions in body frame and use IK to produce target joint angles.
        \item Use a phase evolving in time.
        \item Apply joint position targets using lower-level PD controller with gain 50 and damping .5.
        \item Print debugging metrics you require for iteration (feet contacts, speed, etc.).
        \item Target movement speed is between \textit{<target speed>} and \textit{<target speed>}.
        \item \textit{<gait type>} gait, with foot placement \textit{<feet placement schedule>}
        \item Default feet position for stable standing is directly under the hips.
        \item Environment dt is \SI{2}{\milli\second} with control decimation of 4
        \item keep base roughly horizontal.
        \item Parameterize the gait using parameters from below.
    \end{itemize}

    \vspace{0.5em}
    \textbf{Key Parameters:} gait freq., step length, step height, feet phase offsets, feet default pos. rel. to hip.
\end{promptbox}
\caption{Prompt template we use to generate the gait controllers.}
\label{fig:prompt_example}
\end{figure}

\section{Experimental Details}\label{sec:exp_details}

\textbf{Benchmarks and Datasets.} We evaluate the combination of the action prior and style reward on the benchmark environments from Gymnasium~\cite{towers2024gymnasium} \textit{HalfCheetah-v4}, \textit{Ant-v4}, and \textit{Humanoid-v4}, as well as on the \textit{loco-mujoco} benchmark for \textit{Unitree A1}~\cite{AlHafez.2023} (\autoref{fig:agents}). The source of the demonstration data is different across environments. For \textit{HalfCheetah} and \textit{Ant}, the trajectories are generated by open-loop locomotion controllers~\cite{Raffin.}. For the \textit{Unitree A1}, we use the expert dataset provided by \textit{loco-mujoco}~\cite{AlHafez.2023}. For \textit{Humanoid}, we obtain a demonstration from a pretrained SAC policy~\cite{haarnoja2018softactorcriticoffpolicymaximum}.

The demonstrations consist of a cyclic gait trajectory of 10--212 transitions (\autoref{tab:expert_demos}); no command variation, terrain variation, or perturbation is present in the demonstration data. All demonstrations are collected on flat terrain.

\textbf{Implementation.} For the Gymnasium environments, the policy receives the default observations $o_t$ and tasks rewards $r_{\mathrm{task}}$ provided by the environment. For the quadrupeds, the policy observation consists of the robot proprioceptive state $(q_t,\dot{q}_t)$, the projected gravity vector in the base frame $g_t$, the commanded linear velocity $v_{\mathrm{cmd}}$, the commanded yaw rate $\omega_{\mathrm{cmd}}$, and the previous actions $a_{t-1}, \dots, a_{t-5}$ from the last five control steps. For policies trained with the imitation reward, we additionally provide $\sin(\phi)$ and $\cos(\phi)$ as inputs. Further training parameters are given in~\autoref{tab:training_hps}.

\textbf{Baselines}. We ablate our method (PPO+latent+style) against its individual components of the latent prior (PPO+latent) or a DeepMimic-style reward(PPO+style)~\cite{Peng.2018}. The action prior itself is compared against typical alternative choices, including no prior (PPO), a residual RL setting (PPO+residual+style), and static reference values (PPO+reference+style)~\cite{rudin2022learningwalkminutesusing}. We further add a Behavioral Cloning baseline that imitates the demonstration data.

For torque control, the reference torques were generated by a PD controller that maintains the robot in a standing position. For the residual DRL setting in torque control mode, we use the state-of-the-art approach DecAP~\cite{Sood2023DecAPD}, where an initial position-based policy is decayed towards a torque control policy throughout the training process.

Notably, we do not compare to adversarial-based methods such as AMP~\cite{Peng_2021}, since they require significantly more samples compared to standard PPO (100 million vs our 2 million steps~\cite{Peng_2021}), which is not the intention of our method. Instead, we use DeepMimic-style rewards~\cite{Peng.2018}, which solve a similar problem through explicitly encoded style rewards.

\begin{figure*}[!t]
\vspace{2mm}
  \centering%
  \includegraphics[width=0.2\textwidth]{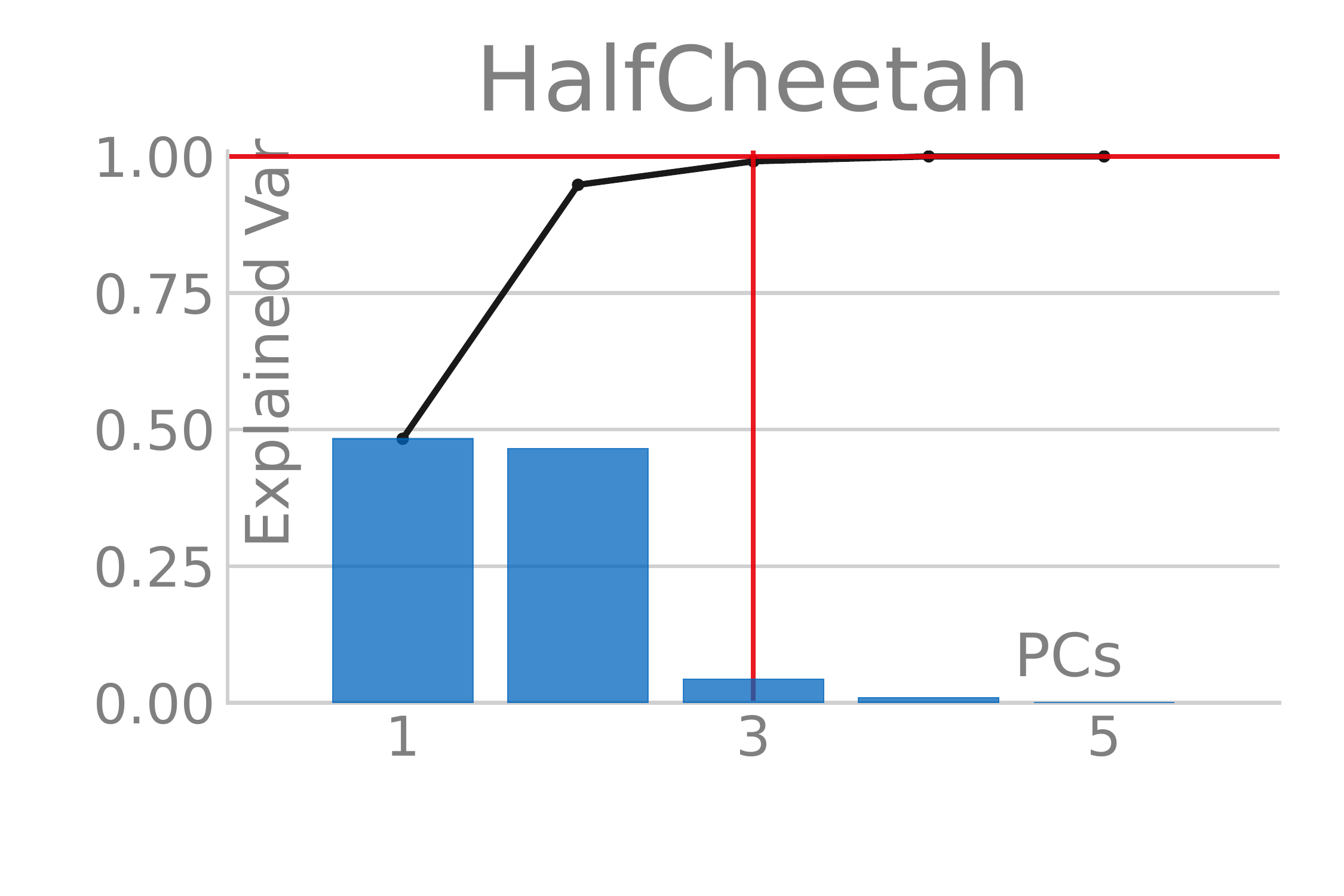}%
\includegraphics[width=0.2\textwidth]{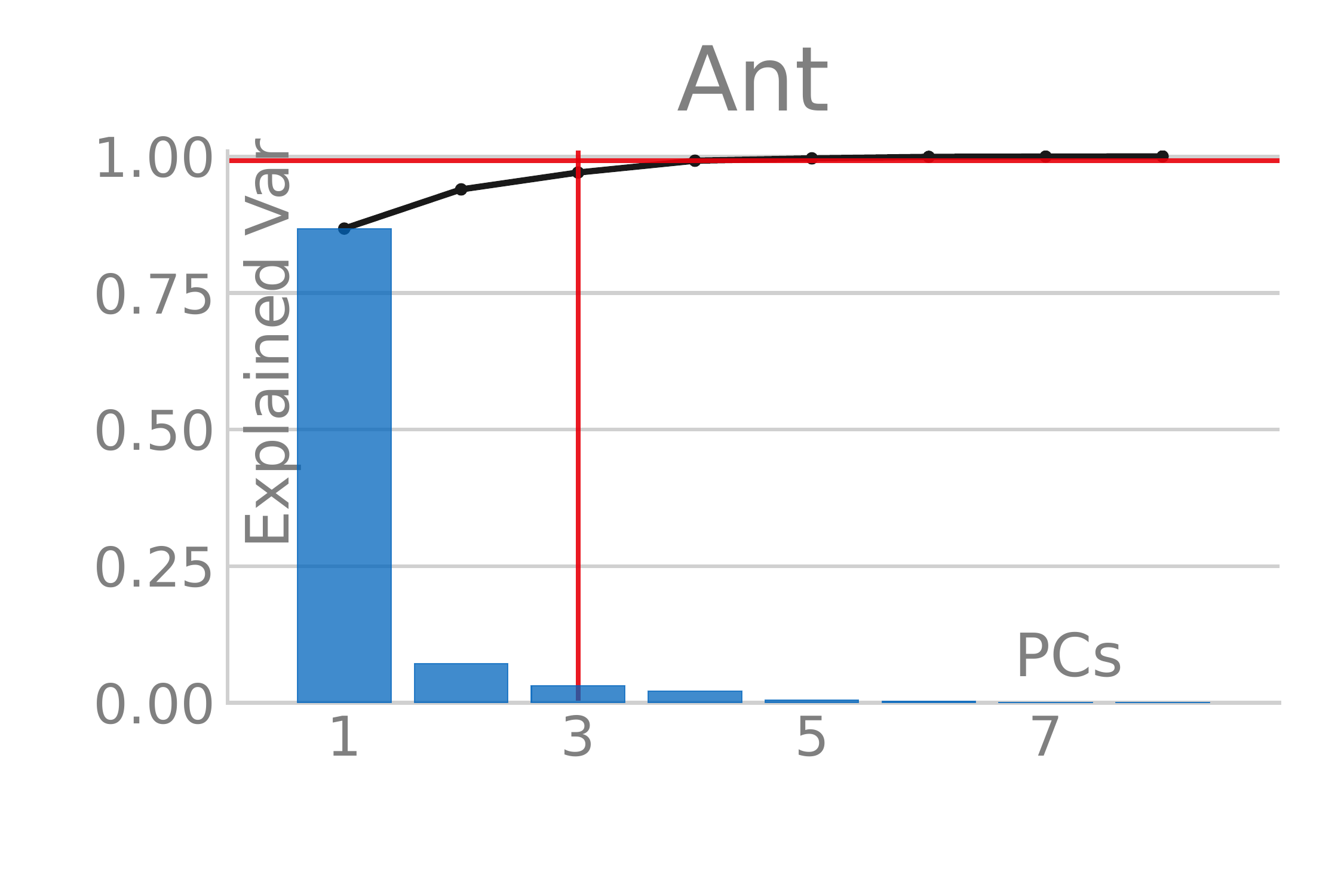}%
\includegraphics[width=0.2\textwidth]{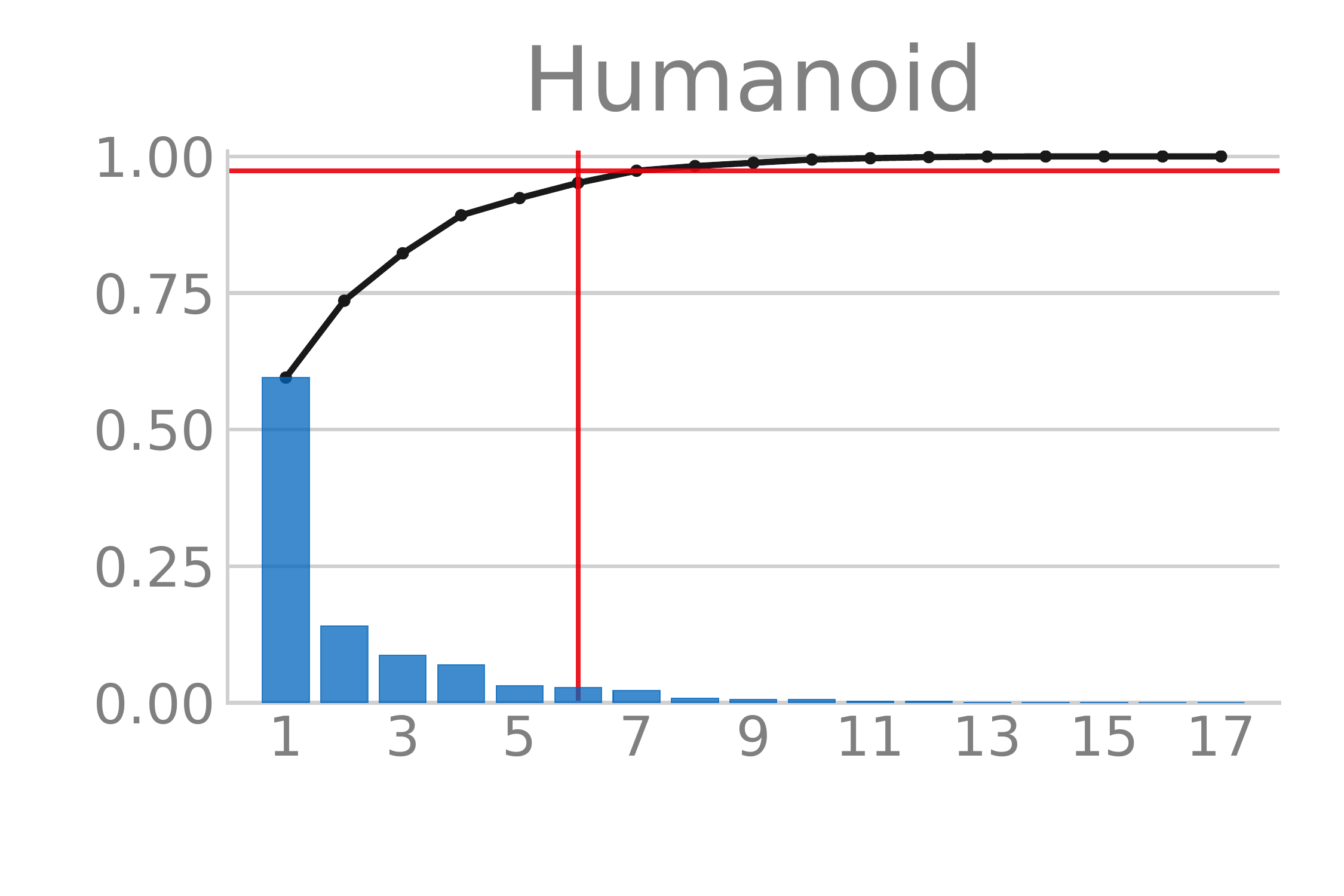}%
\includegraphics[width=0.2\textwidth]{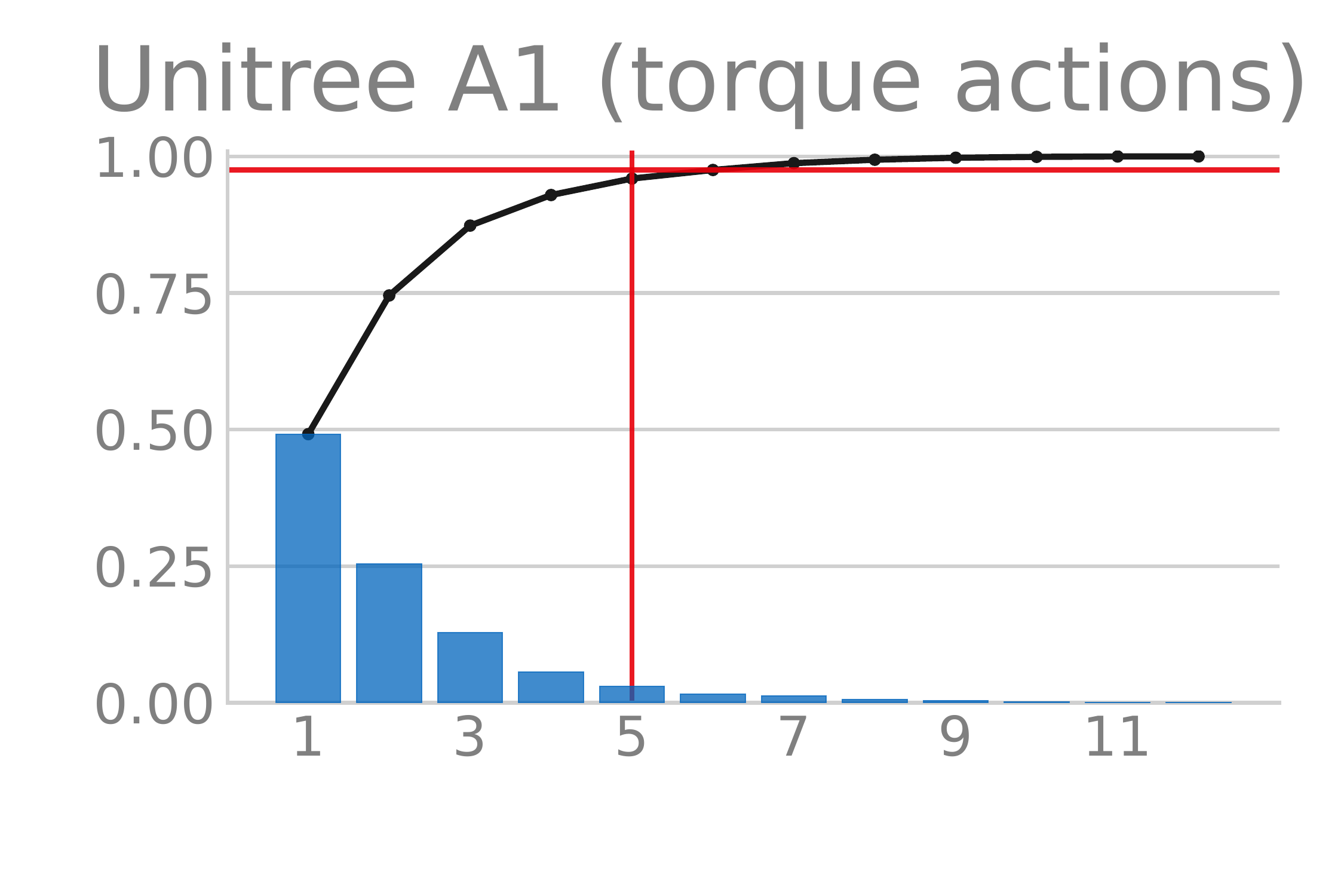}%
\includegraphics[width=0.2\textwidth]{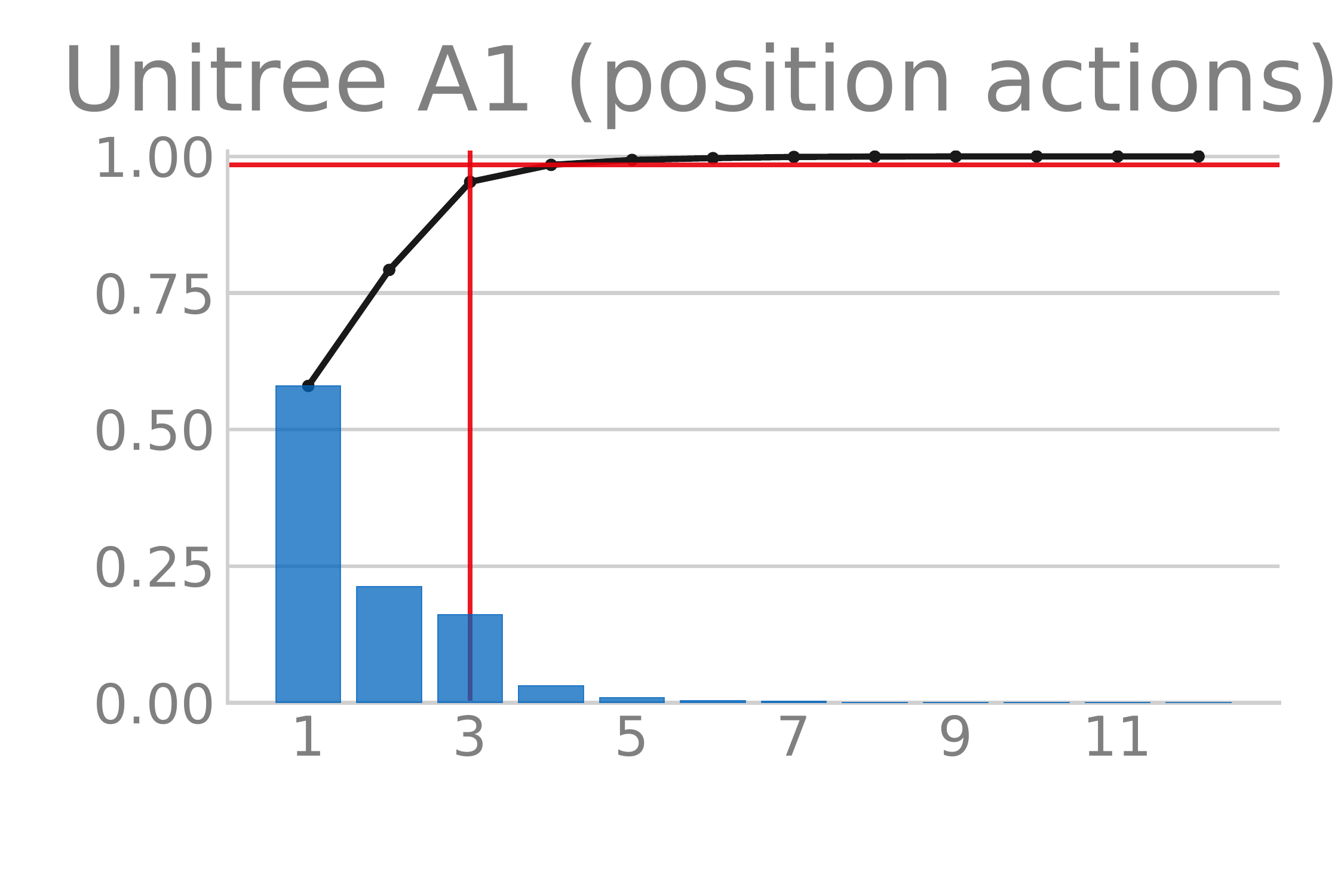}%
  \caption{All motions from the demonstrations can be expressed with only a few Principal components (PCs), motivating the use of a compressed action space that exploits such correlations. PCs left of the vertical red line explain $>$ \SI{95}{\percent} of the variance in the actions and the black lines show the cumulative explained variance. Results from data in \autoref{tab:expert_demos}.}%
  \label{fig:pcas}%
\end{figure*}%

\begin{figure*}[!t]
  \centering%
\includegraphics[width=1\textwidth]{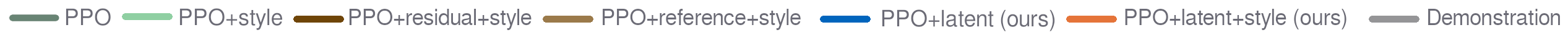}\par%
 \subfloat[]
 {\includegraphics[width=0.2\textwidth]{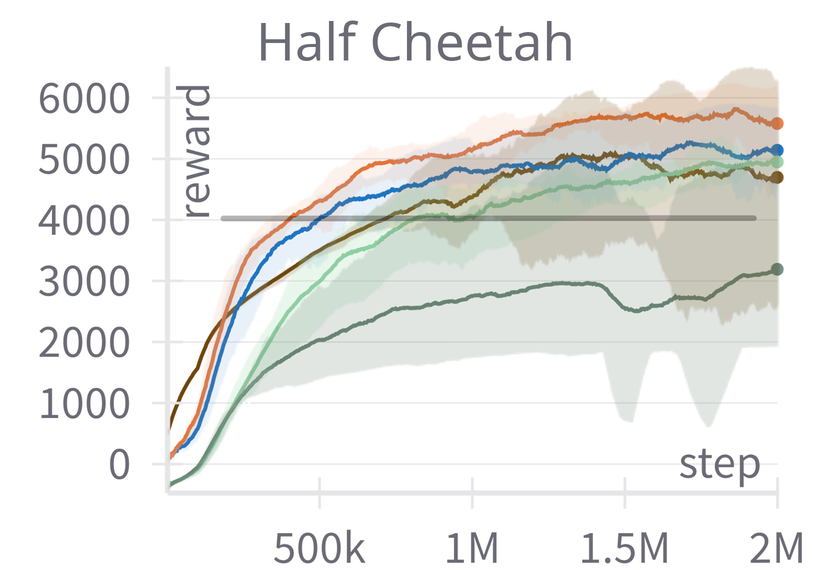}}%
  \subfloat[]{\includegraphics[width=0.2\textwidth]{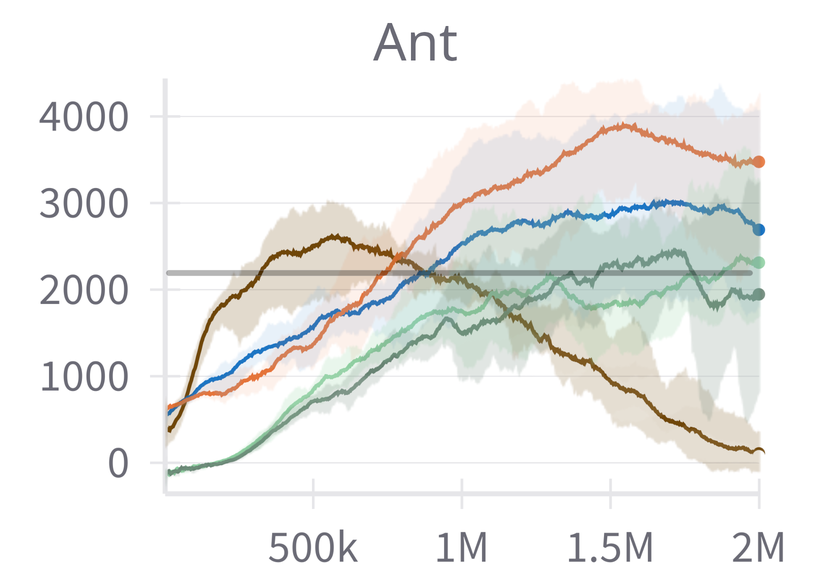}}%
  \subfloat[]{\includegraphics[width=0.2\textwidth]{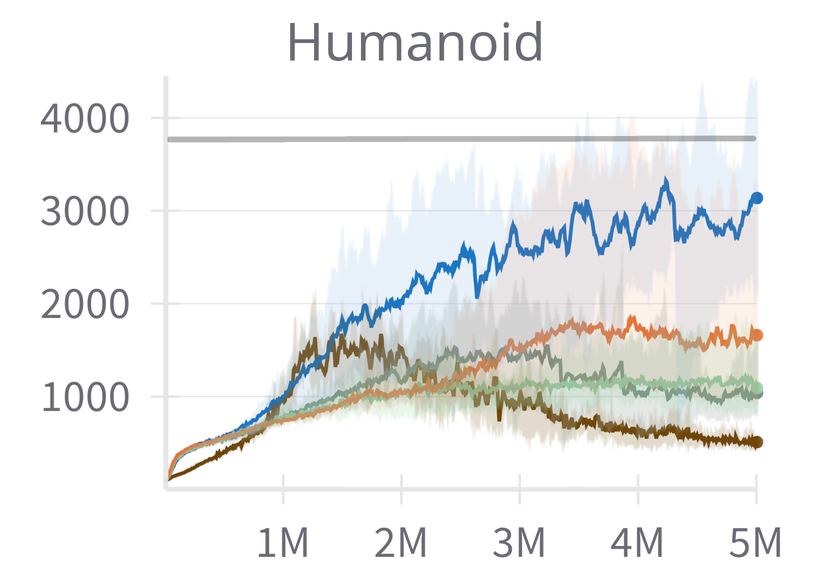}}%
  \subfloat[]{\includegraphics[width=0.2\textwidth]{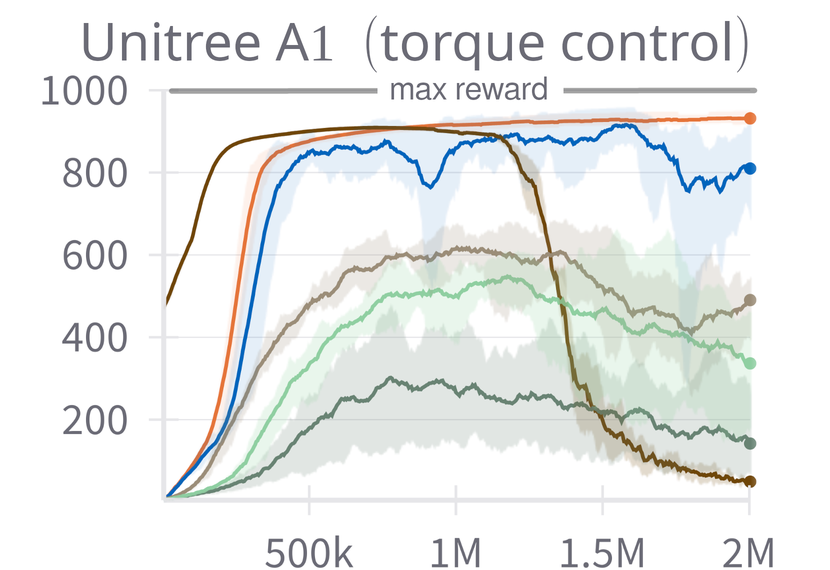}}%
  \subfloat[]{\includegraphics[width=0.2\textwidth]{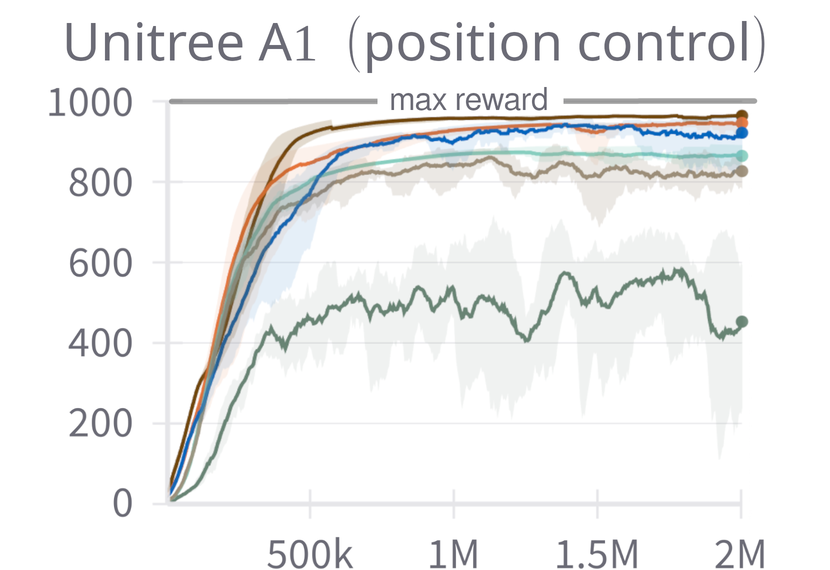}}%
  \caption{Results on benchmark environments. A data-driven action prior significantly outperforms using unbiased PPO and a simple reference position in the action space regarding both sample efficiency during DRL and maximum reward for all environments. This strongly motivates the use of demonstrations for policy learning rather than static joint position references. Adding a single style reward term (+style) creates synergistic effects except for Humanoid, where the demonstration is brittle. We run all experiments on five seeds and report the means and standard deviations on task rewards. BC is omitted as it did not improve upon PPO.}  %
  \label{fig:results}%
\end{figure*}%

\section{Results}\label{sec:results}

\textbf{1) Does a latent action prior trained on expert data improve policy learning?}
We first verify the premise underlying the action prior: that demonstrated actions occupy a low-dimensional subspace. A principal component analysis across all environments confirms this (\autoref{fig:pcas}); a latent dimension of half the full action space captures more than \SI{95}{\percent} of the variance in every case. We therefore set $\dim(a_l) = 0.5\,\dim(a_{full})$ throughout unless stated otherwise.

The action decoder is a shallow network containing, for instance, 240 trainable parameters for the Unitree A1, 0.25\% of the policy. Training on the 10-212 available data points takes only minutes on a single workstation GPU. This is a negligible computation compared with DRL training, and is not reported in the following results figures.

We compare three action-space priors under an identical single style reward: a static offset toward a standing configuration, a residual policy, and our latent prior (\autoref{fig:results}). The latent prior improves both sample efficiency (samples until convergence in DRL) and asymptotic task reward over unbiased PPO and the static offset across all environments, indicating that the agent exploits the correlational structure encoded in the pretrained decoder. On \emph{HalfCheetah} and \emph{Ant}, performance is not bounded by that of the demonstration: the agent surpasses demonstration-level reward, showing that the decoder constrains the action space without capping task performance.

The static offset (PPO+reference+style) should in principle assist learning, since PPO explores around a near-zero-centered mean~\cite{Andrychowicz.2020}. For position control this yields a clear gain in both sample efficiency and final reward. Under torque control the benefit largely disappears, suggesting that torque control is more difficult to learn even when desired offsets for the torques are known. In contrast to position control, even minor deviations from the reference torques can cause the robot to slip and fall, making learning more difficult.

Residual learning (PPO+residual+style) gives mixed results. It helps in several environments, but residual DRL constrains the agent close to the demonstration, limiting reactivity. For torque control we follow DecAP~\cite{Sood2023DecAPD} and decay an initial position-based policy over training. However, the decay of the position control policy alters the environment. This distribution shift is challenging to handle for the DRL agent, which ultimately leads to policy collapse. Compared to a decaying policy, our latent action priors do not introduce changes in the environment, enabling efficient learning also for fewer training samples.%

\begin{figure}[!t]
\vspace{2mm}
  \centering%
\includegraphics[width=0.5\columnwidth]{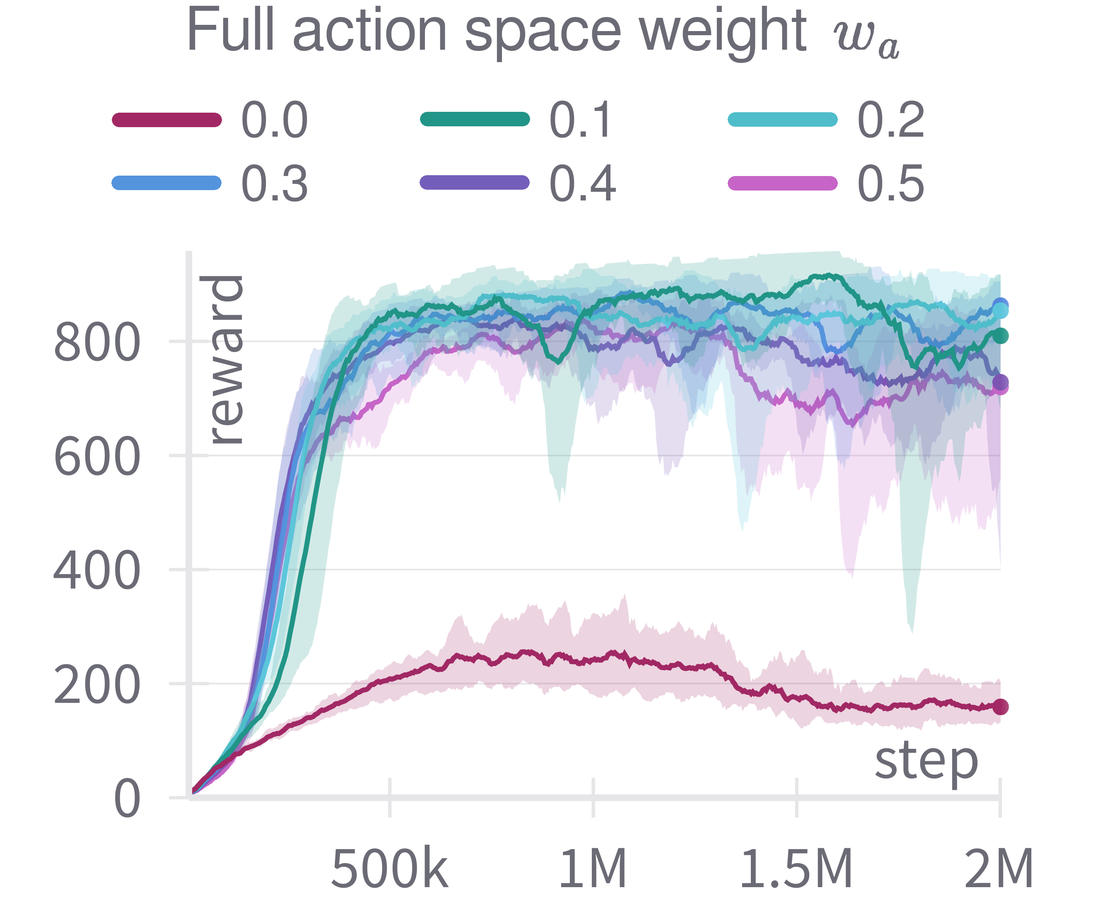}%
\includegraphics[width=0.5\columnwidth]{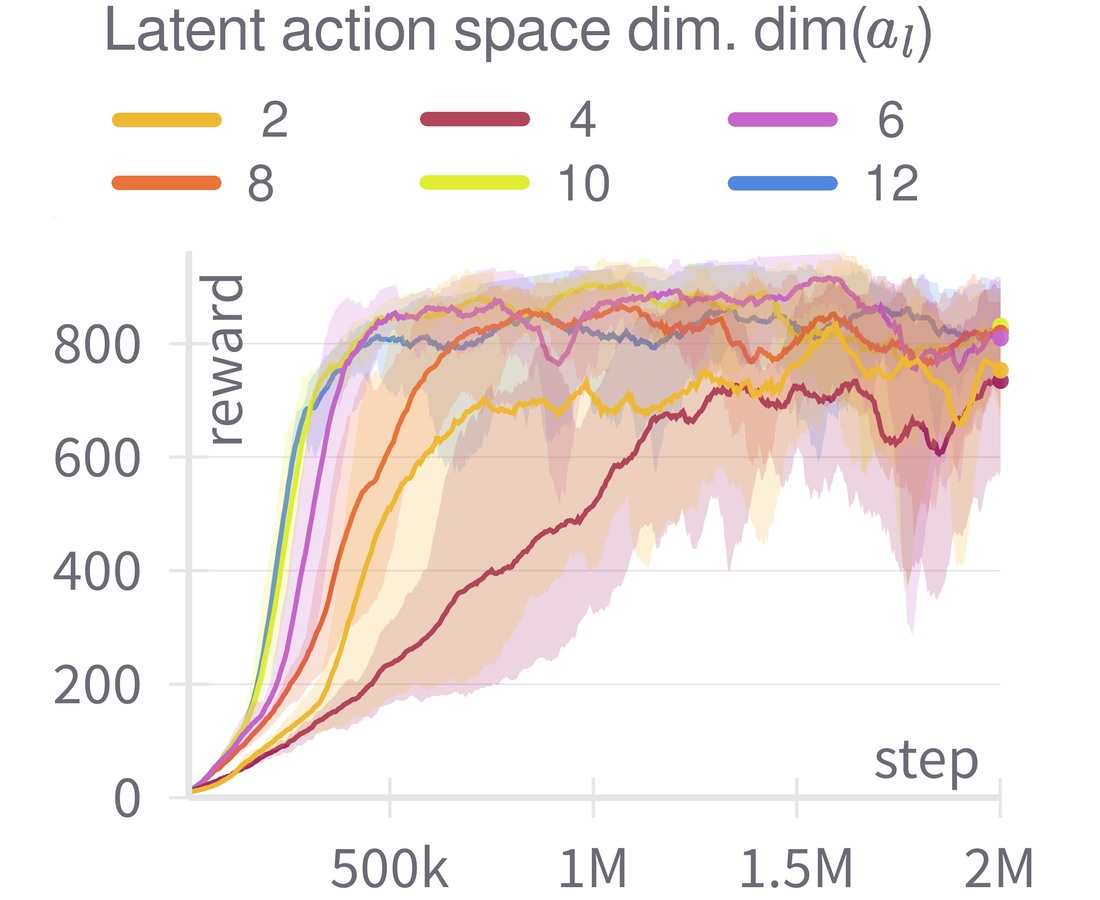}%
  \caption{Policy learning for different hyperparameters introduced by the latent action prior: action space weighting $w_a$ and the dimension of the latent action space dim($a_l$) for the \emph{Unitree A1}. Shown are the mean and standard deviation on 5 seeds.}%
  \label{fig:results_sensitivity}%
\end{figure}

\textbf{Sensitivity to action-space parameters.}
A central claim is that the introduced hyperparameters require comparatively little tuning effort. To this end, we investigate the sensitivity of our approach towards the two parameters introduced --- the latent dimension $\dim(a_l)$ and the residual weight $w_a$ --- on the \emph{Unitree A1} (\autoref{fig:results_sensitivity}). Fast convergence and high task reward are obtained across a broad range of both. Even at $w_a = 0.5$, where the decoded latent action accounts for only half of the applied command, the agent still benefits from the prior. Setting the residual action space to zero ($w_a=0.0$) hinders learning, as the agent starts in a random state and experiences perturbations. To recover stable locomotion, the agent must take actions beyond the demonstrated distribution. We recommend a full action space weight $w_a$ in $[0.1, 0.5]$, where unstable agents should use higher values (\autoref{tab:training_hps}).

For the latent dimension, we find --- contrary to our initial expectation --- that larger $\dim(a_l)$ performs on par with or better than aggressive compression. Even at $\dim(a_l) = \dim(a_{full})$ ($12$ for the \emph{Unitree A1}), learning improves markedly over the unbiased baseline.  This indicates that not the dimensionality reduction, but the information about the correlations in the expert's actions improves exploration and policy learning.

\begin{figure}[!t]
\vspace{2mm}
    \centering
\includegraphics[width=1\columnwidth]{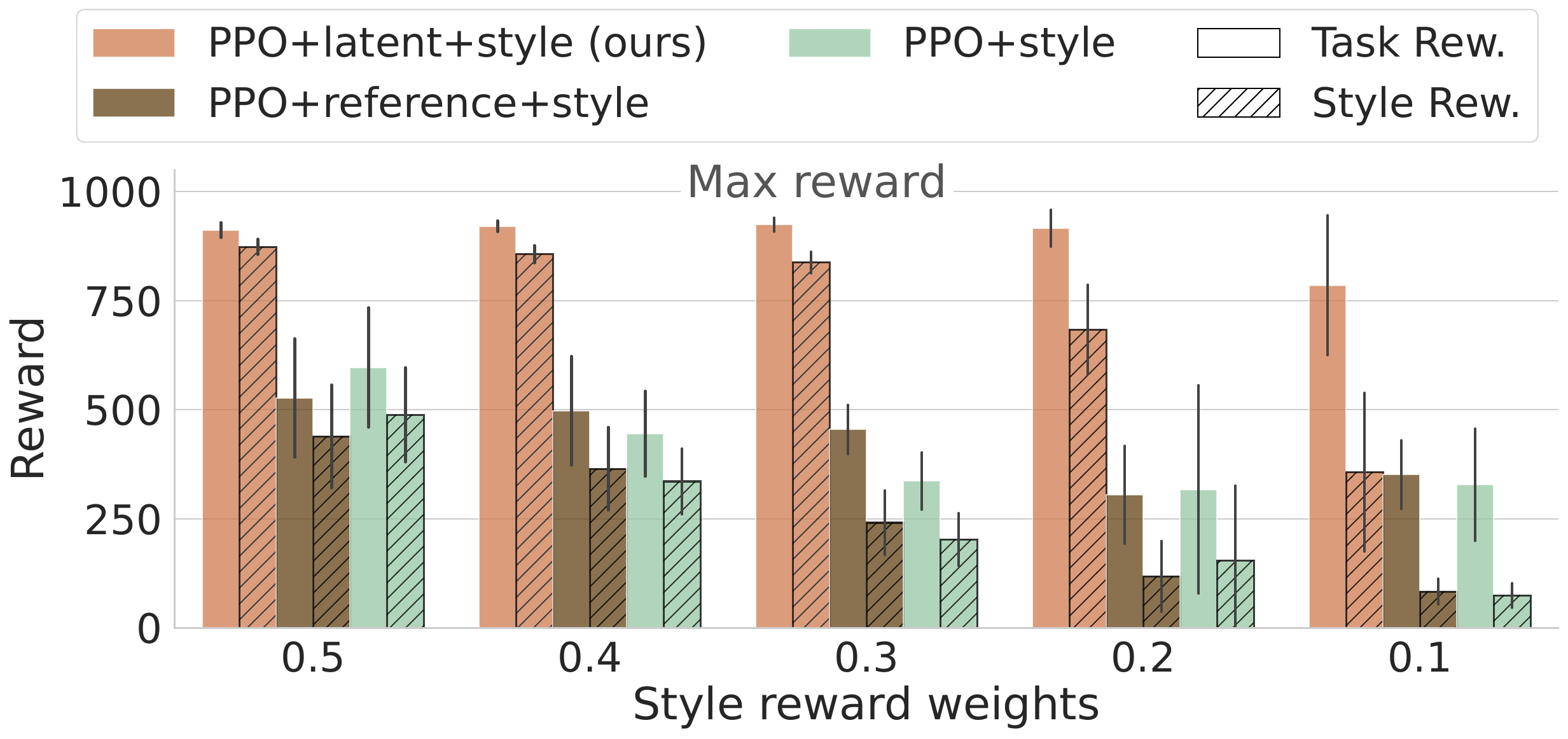}
    \caption{The combination of a latent action prior with a single style reward term is significantly less sensitive towards reward tuning than baselines with only DeepMimic-style style rewards and reference joint positions. Shown is the ablation on the style reward weights $w_{style}$ for \emph{Unitree A1} ($w_{\mathrm{task}} = 1 -w_{\mathrm{style}}$). Shown are the mean and standard deviation on 5 seeds.}
    \label{fig:results_style_reward_ablations}
\end{figure}

\textbf{2) Does a single style reward term improve the resulting gaits, and is it sensitive to parameter tuning?}
While each prior improves imitation on its own, their combination is what yields robust reproduction of the demonstrated gait (\autoref{fig:results_style_reward_ablations}, \autoref{fig:results}).

Adding the style reward of \autoref{eq:style} to the latent prior raises task reward in most environments and produces consistent imitation across the full sweep of reward weights $w_{style} \in [0.1, 0.5]$ (\autoref{fig:results_style_reward_ablations}). Baselines relying on the style reward alone or on a static reference degrade sharply as $w_{style}$ decreases. We attribute this robustness to a mutually reinforcing feedback-loop: the decoder biases exploration toward demonstrated actions, which raises the style reward, which in turn favors policies operating within the demonstrated manifold.

\begin{figure}[t]
    \centering
    \vspace{2mm}
\includegraphics[width=1\columnwidth]{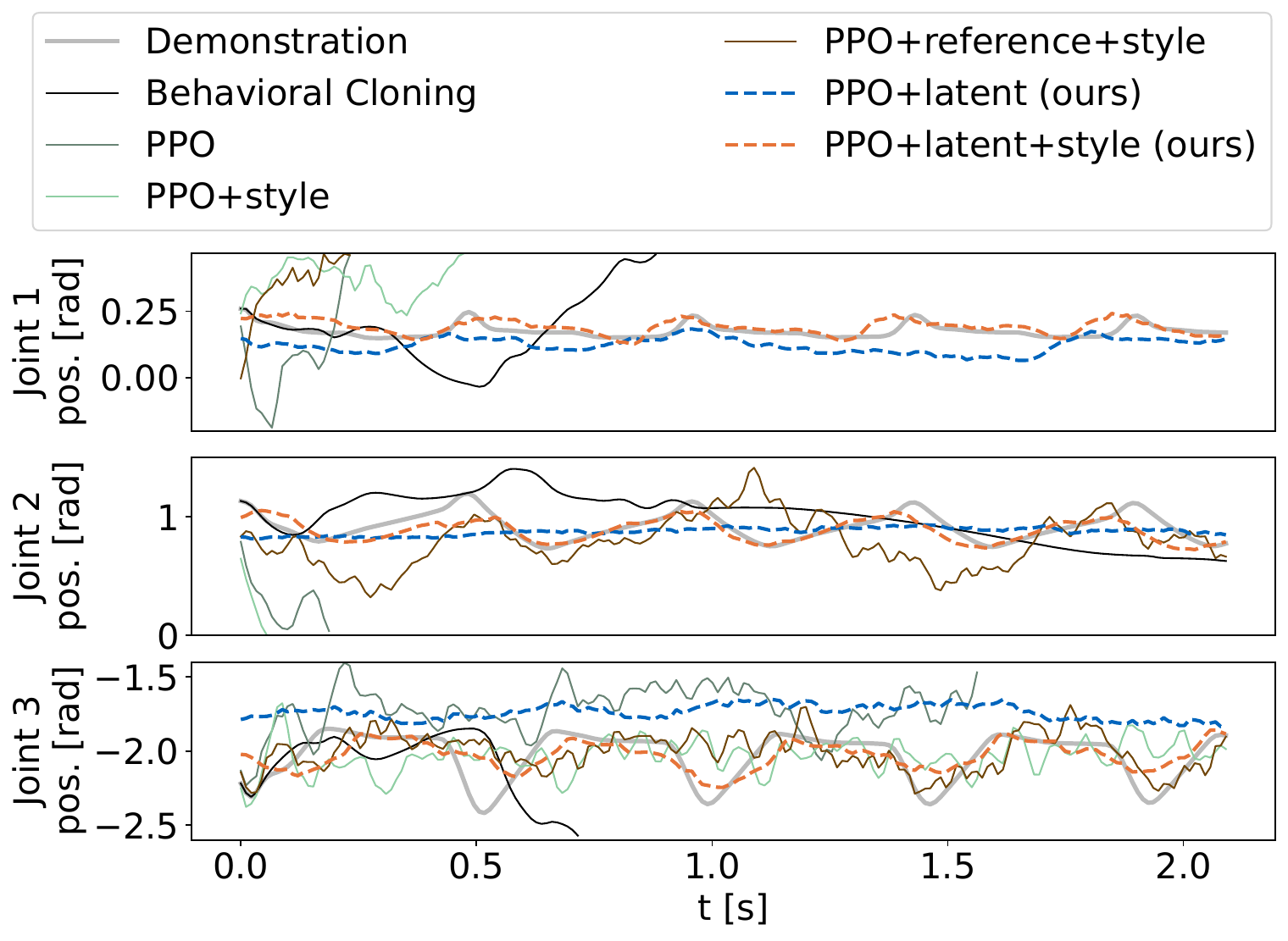}
    \caption{Joint positions for the left hind leg of the Unitree A1 for torque control in simulation. The synergy of both priors leads to joint positions close to the expert. The BC policy tracks the expert demonstration for the first frames but diverges quickly. RMSE [rad]: PPO+latent+style: 0.05, PPO+latent: 0.17, PPO+reference+style: 0.31, PPO+style: 0.52.}
    \label{fig:results_jpos}
\end{figure}

\autoref{fig:results_jpos} presents this result qualitatively. The action prior alone (PPO+latent) recovers joint configurations close to the demonstration (RMSE \SI{0.17}{\radian}) but exhibits high-frequency oscillation around static configuration, since it only constrains the actions spatially. Adding the style reward provides additional temporal alignment and reduces the tracking error to \SI{0.05}{\radian}, against \SI{0.31}{\radian} for a static reference and \SI{0.52}{\radian} for the style reward alone. Behavioral cloning tracks the demonstration for the initial frames before diverging, as expected from compounding out-of-distribution errors. More qualitative results are contained in the supplementary video.

The one exception is \emph{Humanoid}, where the style reward does not improve on the action prior. Here, the demonstration originates from an unregularized SAC policy that itself exhibits brittle, unnatural gait. We hypothesize that the demonstration is a local optimum that is difficult to replicate and has limited value. The combination of a high-dimensional agent and a pure forward-velocity task reward further dilutes the benefit --- suggesting that the quality of the demonstration determines the strength of the synergy.

\begin{figure}[!t]
  \centering%
  \subfloat[]{\includegraphics[width=0.32\columnwidth]{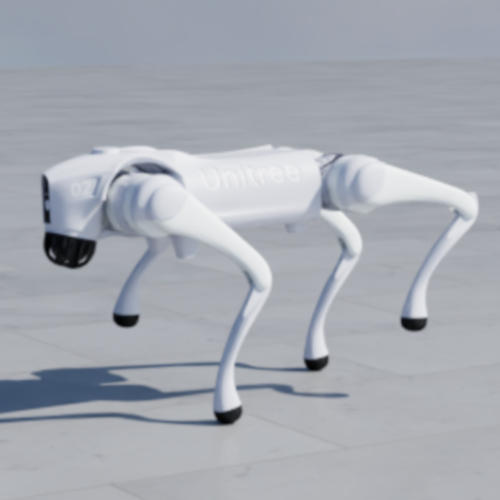}}%
  \hspace{0.012\columnwidth}%
  \subfloat[]{\includegraphics[width=0.32\columnwidth]{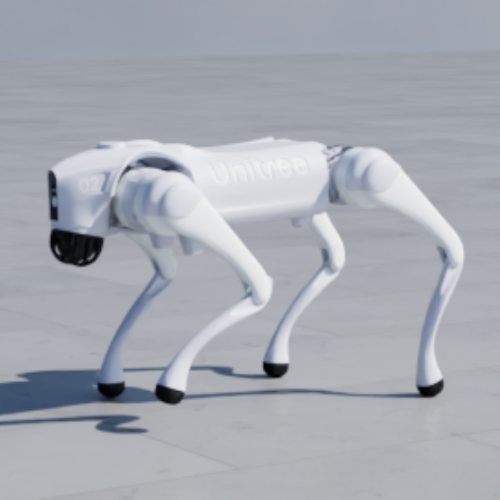}}%
  \hspace{0.012\columnwidth}%
  \subfloat[]{\includegraphics[width=0.32\columnwidth]{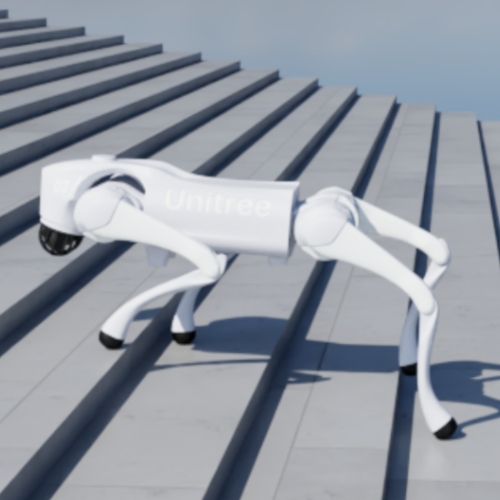}}\\
  \vspace{-0.035\columnwidth}
  \subfloat[]{\includegraphics[width=0.95\columnwidth]{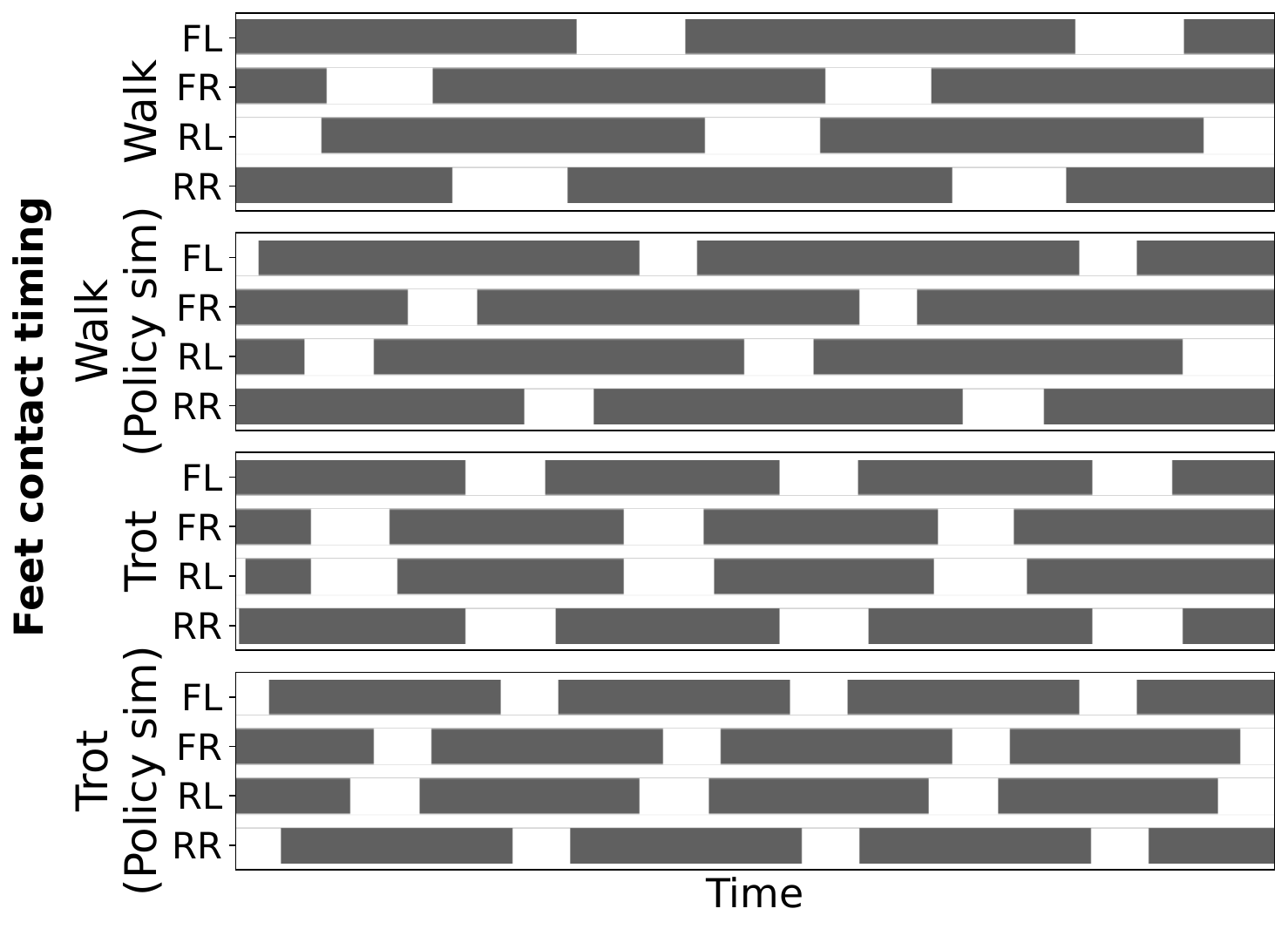}}
  \caption{LLMs can effectively design open-loop gait controllers for various locomotion scenarios leveraging established methods~\cite{zeng2019legtrajectory,iscen2018pmtg}: (a) trot, (b) walk, and (c) stair climbing. Fig. (d) Shows the produced feet contact timing for the trotting and walking gaits when executing the open-loop controllers in simulation and the resulting policies. The feet contact patterns translate to the trained policy indicating imitation.}%
  \label{fig:gaits}%
\end{figure}

\textbf{3) + \textbf{4)} Can an LLM design the required demonstrations, and can the resulting policies be deployed on hardware?}
For the \emph{Unitree Go2}, we applied the LLM-guided procedure described in \autoref{sec:method} to three locomotion scenarios: walking, trotting, and stair climbing (\autoref{fig:gaits}). The LLM (Claude 3 Opus) iterates by executing the controller in simulation and inspecting the printed debugging metrics until the requirements are met. A gait converges in 10 prompt–simulate cycles and two human interventions on average (30–45 \si{\min} per single gait). The outcome of the reference gait can immediately be observed in simulation rather than after full DRL training, and is verified by a human. After generating the demonstrations, we train the downstream policies and deploy the results to hardware in both position- and torque-control modes for trotting gait.

The LLM produces the desired gaits for trotting, walking, and stair climbing as shown by the corresponding foot contact patterns (\autoref{fig:gaits}). These schedules are inherited by the downstream trained and deployed policies, showing that the action prior and style reward provide structural information to constrain policy learning.

\textbf{Training.}~The control frequency during training is \SI{50}{\hertz} for position control and \SI{500}{\hertz} for torque control, with simulation decimations of 4 and 2. To support sim-to-real transfer we randomize friction coefficients, PD gains, base mass, center of mass, gravity, and action delays (up to 4 control steps), and apply pushes of up to \SI{1}{\metre\per\second}, adopting the ranges of~\cite{bohlinger2024policyrunallendtoend}. All remaining hyperparameters follow the \emph{Unitree A1} configuration. Training takes roughly one hour on a single NVIDIA RTX 4090.

\textbf{Deployment.}~Policies are converted to C++ with runtime optimizations to meet the \SI{500}{\hertz} torque-control target: BFloat16 quantization~\cite{jacob2017quantizationtrainingneuralnetworks}, relaxed floating-point operations for compiler-level optimization of function evaluations (\texttt{-ffast-math})~\cite{dongarra2024hardwaretrendsimpactingfloatingpoint}, and just-in-time compilation via TorchScript~\cite{paszke2019pytorch} to increase inference speed. The torque control policy runs at \SI{500}{\Hz}, while \SI{50}{\Hz} is sufficient for position control, both on the CPU of an external laptop. The Ethernet network delay is \SI{0.5}{\milli\second}, which is sufficient for low-latency torque control. 

\begin{figure}[t]
  \centering%
  \vspace{2mm}
\includegraphics[width=1.0\columnwidth]{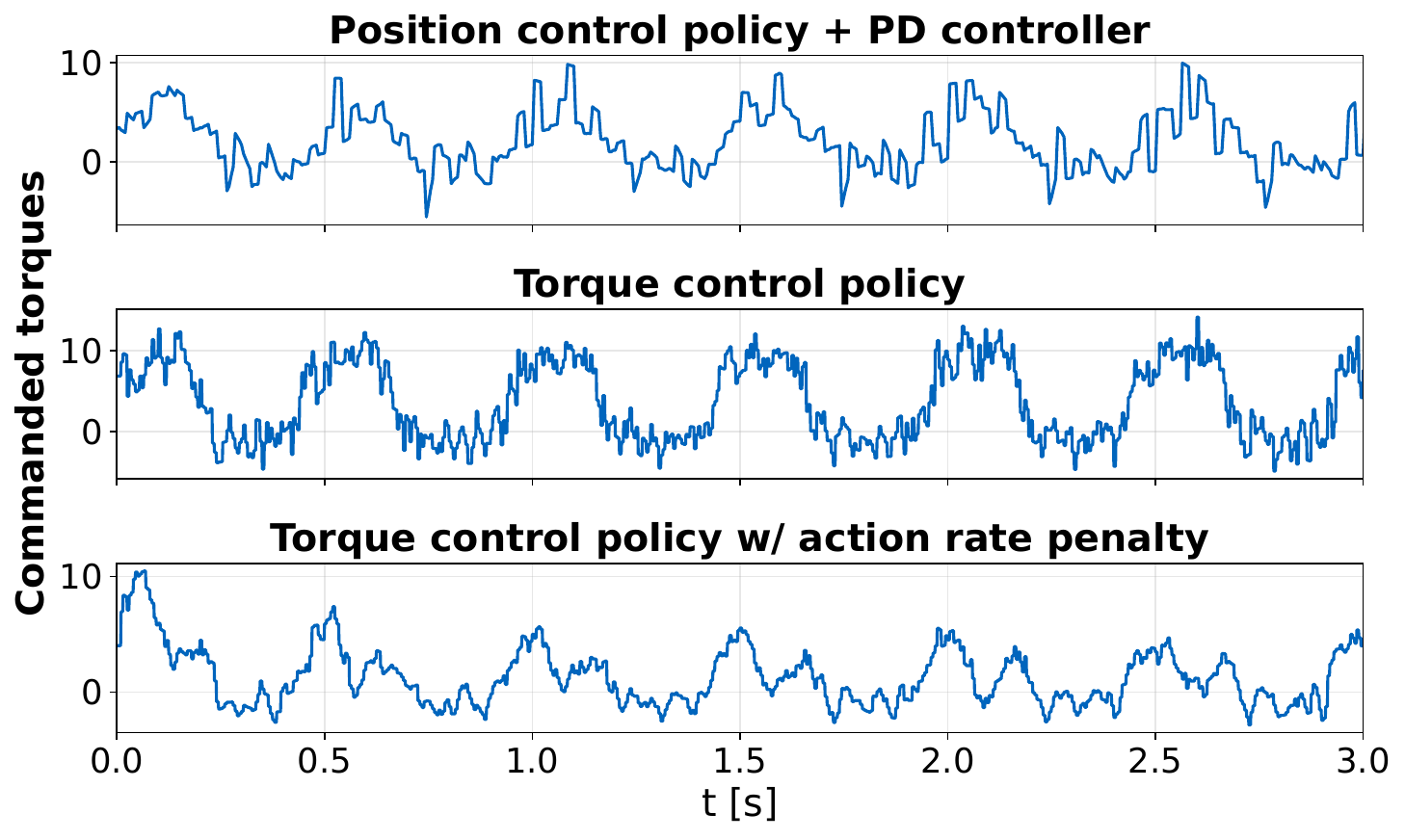}%
  \caption{While position control can directly be deployed to hardware, torque control on the robot leads to high-frequency vibrations when no penalty for the action rate is added. Shown are the commanded torques sent to the robot's rear left calf joint.}%
  \label{fig:torques}%
\end{figure}

We  deploy the trotting policy to hardware for both position and direct \SI{500}{\hertz} torque control (see supplementary video). Despite demonstrating only a single trotting motion, we obtain steerable behaviors for both indoor and outdoor environments. The policies are robust to  external disturbances and can cross rough terrains. For direct torque control mode, in practice we find that one additional reward term is needed for deployment. This term penalizes the policy's action rate; otherwise, we observe high-frequency vibrations on the robot. Adding this action rate penalty in addition to the style reward is key to avoiding high-frequency vibrations. We compare deployment with and without the reward term in \autoref{fig:torques} and the supplementary video. Overall, the results show that the combination of action and style prior provides sufficient structural information to deploy robust policies that imitate a given demonstration to hardware.

\section{Discussion}
Our results support a simple hypothesis: the action prior accelerates learning and raises task reward, while the style reward supplies the temporal structure that renders the motion visually plausible and deployable. Neither is sufficient alone, and the robustness to reward weighting is a property of the combination (\autoref{fig:results_style_reward_ablations}). We further show that LLMs can aid in generating the required demonstrations using established methods for open-loop locomotion control.

Torque control is notoriously difficult to train and deploy, yet offers robustness, compliance, and dynamic capability~\cite{Sood2023DecAPD, chen2023learningtorquecontrolquadrupedal}. Our method provides an effective inductive bias even under challenging torque control. We find that one additional reward term for limiting the action rate change is necessary for deployment. This, however, reduced tuning effort compared to previous methods that rely on additional reward terms or introduce brittle hyperparameters.

\textbf{Limitations.} Our demonstrations are cyclic and single-task. The phase variable required by the style reward further presumes a periodic reference. As the decoder spans only demonstrated actions, we expect generalization to nearby behaviors (varying commands, perturbations, rough terrain), but not to non-periodic ones such as jumping or fall recovery. The benefit also depends on demonstration quality and embodiment (Humanoid, \autoref{fig:results}). Extending the formulation to multi-task or aperiodic behavior requires replacing the phase-indexed reward with formulations from FLD~\cite{li2024fld} or AMP~\cite{Peng_2021}. On the prior side, a Gaussian mixture~\cite{viroli2017deepgaussianmixturemodels} or variational model could increase the capacity of the latent action space to cover diverse behaviors.

\section{Conclusion}
Obtaining demonstrations and deriving suitable priors without extensive reward tuning remains challenging for learning locomotion. We presented a method for learning deployable locomotion from a small state--action demonstration set using a simple action prior and style reward term. Crucially, we show that the required demonstrations can be designed by an LLM through a simulation-in-the-loop procedure, and the resulting policies transfer to hardware for both position and direct \SI{500}{\hertz} torque control. This constitutes a pipeline from natural-language specification to low-level joint torques, and contributes to a broader shift in robot learning --- from hand-engineered priors toward data-driven priors elicited from foundation models.

\begin{table}[!h]
\scriptsize
\captionsetup{skip=2pt}
\caption{Training parameters.}
\label{tab:training_hps}
\centering
\begin{threeparttable} %
\def\arraystretch{1.2}%
\begin{tabularx}{\columnwidth}{|l|c|C|c|c|}
\hline
\multicolumn{1}{|c|}{\textbf{Env}} & $\textbf{dim}\bm{(a_l)}$ & $\bm{w_a}$ & \textbf{Samples / Update} & \textbf{Policy} \\ \hline
HC-v4 & 0.5 $\dim(a_{full})$ & 0.1 & 2048 & [64x64] \\ \hline
Ant-v4 & 0.5 $\dim(a_{full})$ & 0.1 & 2048 & [64x64] \\ \hline
UnitreeA1/Go2 & 0.5 $\dim(a_{full})$ & 0.1 & 2x2048 & [256,256]  \\ \hline
Humanoid-v4 & 0.5 $\dim(a_{full})$ & 0.5 & 4x2048 & [512x512] \\ \hline
\end{tabularx}
\begin{tablenotes}
\end{tablenotes}
\end{threeparttable}
\end{table}

{\setlength{\intextsep}{4pt}
\begin{table}[!h]
\scriptsize
\captionsetup{skip=2pt}
\caption{Expert demonstrations.}
\label{tab:expert_demos}
\centering
\begin{threeparttable} %
\def\arraystretch{1.2}%
\begin{tabularx}{\columnwidth}{|l|C|X|}
\hline
\multicolumn{1}{|c|}{\textbf{Env}} & \textbf{Frames}\tnote{a} & \multicolumn{1}{c|}{\textbf{Source}} \\ \hline
HalfCheetah-v4 & 10  & Open-loop oscillator \cite{Raffin.} \\ \hline
Ant-v4         & 22  & Open-loop oscillator \cite{Raffin.} \\ \hline
Humanoid-v4    & 212  & Off-policy DRL \cite{Kuznetsov.2020} \\ \hline
Unitree A1     & 92  & loco-mujoco \cite{AlHafez.2023} \\ \hline
Unitree Go2     & 112--120  & Open-loop controller (LLM) \\ \hline
\end{tabularx}
\begin{tablenotes}
\scriptsize
\item[a] State--action pairs of a cyclic gait used in the demonstration data set.
\end{tablenotes}
\end{threeparttable}
\end{table}

\balance
\renewcommand*{\bibfont}{\small}
\printbibliography

\end{document}